\documentclass[letterpaper]{article} % DO NOT CHANGE THIS
\usepackage[draft]{aaai2026}  % DO NOT CHANGE THIS

\usepackage{times}  % DO NOT CHANGE THIS
\usepackage{helvet}  % DO NOT CHANGE THIS
\usepackage{courier}  % DO NOT CHANGE THIS
\usepackage[hyphens]{url}  % DO NOT CHANGE THIS
\usepackage{graphicx} % DO NOT CHANGE THIS
\usepackage{natbib}  % DO NOT CHANGE THIS AND DO NOT ADD ANY OPTIONS TO IT
\usepackage{caption} % DO NOT CHANGE THIS AND DO NOT ADD ANY OPTIONS TO IT
\usepackage{algorithm}
\usepackage{algorithmic}

\usepackage{booktabs} % 提供 professional 线条
\usepackage{multirow} % 提供合并单元格功能
\usepackage{array} % 提供额外的列格式
\usepackage{enumitem}
\usepackage{amsmath}
\usepackage{newfloat}
\usepackage{listings}
\DeclareCaptionStyle{ruled}{labelfont=normalfont,labelsep=colon,strut=off} % DO NOT CHANGE THIS
\floatstyle{ruled}
\newfloat{listing}{tb}{lst}{}
\floatname{listing}{Listing}
\usepackage{amsmath, amssymb, amsfonts}

\title{UI-UG: A Unified MLLM for UI Understanding and Generation}

\author{
    Hao Yang\textsuperscript{\rm 1},
    Weijie Qiu\thanks{These authors contributed equally.}\textsuperscript{\rm 1,2},
    Ru Zhang\footnotemark[1]\textsuperscript{\rm 1,3},
    Zhou Fang\textsuperscript{\rm 1},
    Ruichao Mao\textsuperscript{\rm 1},
    Xiaoyu Lin\textsuperscript{\rm 1},
    Maji Huang\textsuperscript{\rm 1},
    Zhaosong Huang\textsuperscript{\rm 1},
    Teng Guo\textsuperscript{\rm 1},
    Shuoyang Liu\textsuperscript{\rm 1},
    Hai Rao\textsuperscript{\rm 1}
}

\affiliations{
    \textsuperscript{\rm 1}Ant Group \\
    \textsuperscript{\rm 2}Beijing University of Posts and Telecommunications \\
    \textsuperscript{\rm 3}Zhejiang University
}

\usepackage{bibentry}
\begin{document}

\maketitle

\begin{abstract}
% \zhaosong{The automatic generation and understanding of user interfaces (UI) using Multimodal Large Language Models (MLLMs) are two common tasks in application development and evaluation.
% However, existing models face challenges in accurately UI comprehension and high-quality UI generation.}
Although Multimodal Large Language Models (MLLMs) have been widely applied across domains, they are still facing challenges in domain-specific tasks, such as User Interface (UI) understanding accuracy and UI generation quality. In this paper, we introduce \textbf{UI-UG} (a unified MLLM for \textbf{UI} \textbf{U}nderstanding and \textbf{G}eneration), integrating both capabilities. For understanding tasks, we employ Supervised Fine-tuning (SFT) combined with Group Relative Policy Optimization (GRPO) to enhance fine-grained understanding on the modern complex UI data. For generation tasks, we further use Direct Preference Optimization (DPO) to make our model generate human-preferred UIs. In addition, we propose an industrially effective workflow, including the design of an LLM-friendly domain-specific language (DSL), training strategies, rendering processes, and evaluation metrics. In experiments, our model achieves state-of-the-art (SOTA) performance on understanding tasks, outperforming both larger general-purpose MLLMs and similarly-sized UI-specialized models. Our model is also on par with these larger MLLMs in UI generation performance at a fraction of the computational cost. We also demonstrate that integrating understanding and generation tasks can improve accuracy and quality for both tasks. 

\textbf{Code \& Model}: https://github.com/neovateai/UI-UG

\end{abstract}

%     \link{Code}{  https://github.com/neovateai/UI-UG}
%     % \link{Datasets}{https://aaai.org/example/datasets}
%     % \link{Extended version}{https://aaai.org/example/extended-version}

\section{Introduction}
With the advent of the modern mobile Internet era, User Interface (UI) plays an increasingly significant role in our daily lives. Whenever an app page is opened, a new UI is presented to the user, which can significantly influence user experience and key business metrics. Consequently, research in the UI domain has been a focal point for both academia and industry.

Front-end developers can also benefit from research focused on UI understanding and generation tasks in twofold. First, these skills improve development efficiency by enabling automated tools like design-to-code (D2C) and low-code UI construction. Additionally, they support real-time user interactions in AI-powered conversational interfaces, allowing features such as automated interactions (e.g., button clicks) and dynamic UI generation during LLM-chat conversations.

Many general-purpose multimodal large language models (MLLMs), such as GPT-4o \cite{gpt4o}, Claude \cite{Claude}, Gemini \cite{gemini}, and Qwen VL series \cite{Qwen-VL,Qwen2-VL,Qwen2.5-VL} possess certain capabilities in the UI domain, but their accuracy and stability remain suboptimal for domain-specific tasks. Meanwhile, recent works leveraging MLLMs in the UI domain have focused on specified UI tasks. For example, Ferret-UI \cite{ferret-ui} and its successor Ferret-UI2 \cite{ferret-ui2} have developed general-purpose models for UI understanding, such as referring, grounding, and extended their applicability across platforms. Concurrently, studies like DCGen \cite{dcgen}, DeclarUI \cite{declareUI} have optimized UI generation by using multimodal models.
% Models like SeeClick, UI-TARS, and UI-TARS 1.5 specialize in UI intent understanding and click prediction, achieving notable results. 

\begin{figure}[t]
\centering
\includegraphics[width=0.95\columnwidth]{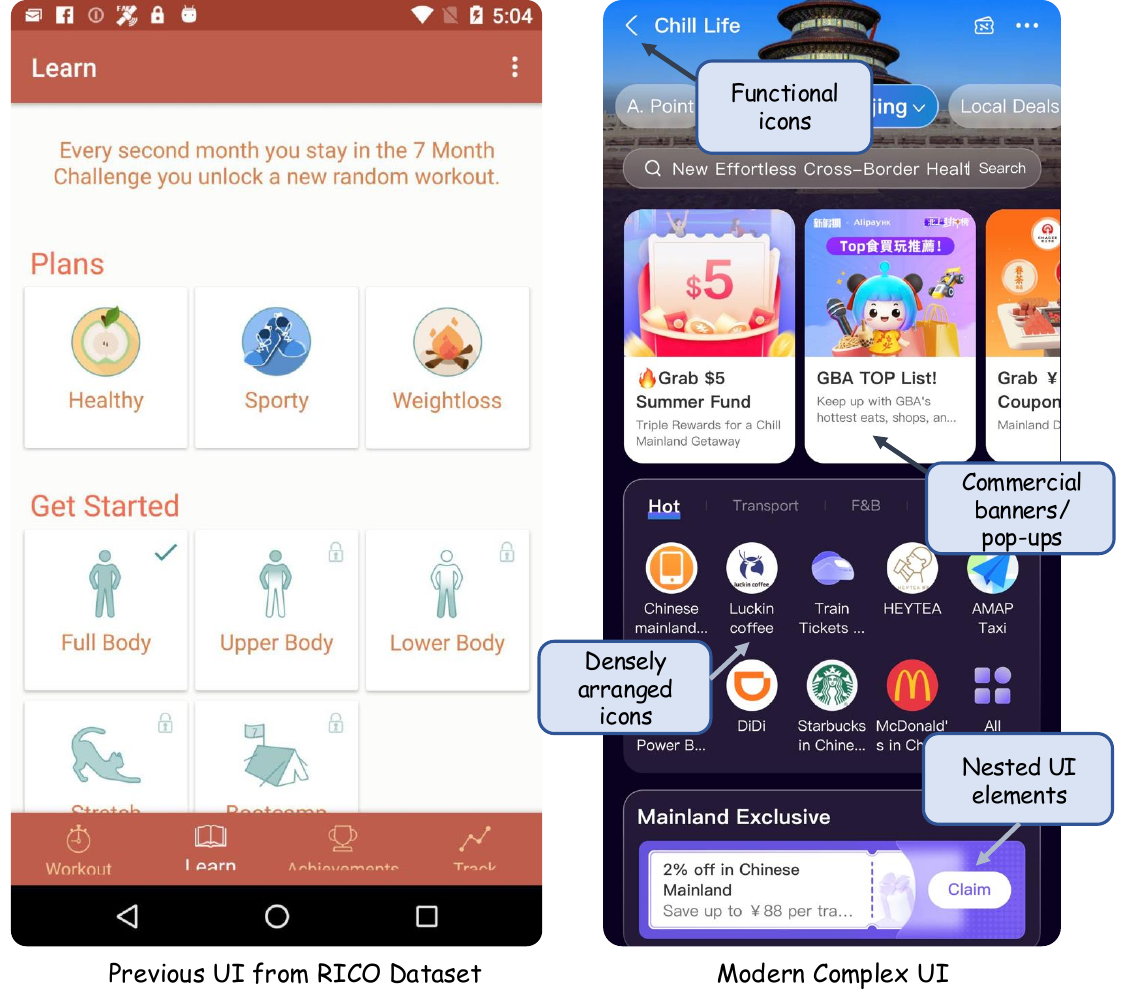} % Reduce the figure size so that it is slightly narrower than the column. Don't use precise values for figure width. This setup will avoid overfull boxes.
\vspace{-0.5em} 
\caption{With the development of UI resolution and design, modern apps now include more icons, ads, and complex elements, creating new challenges in understanding UIs.}
\vspace{-1.3em} 
\label{fig1}
\end{figure}

\begin{figure*}[t]
\centering
% \hspace*{-0.3cm}
\includegraphics[width=2.0\columnwidth]{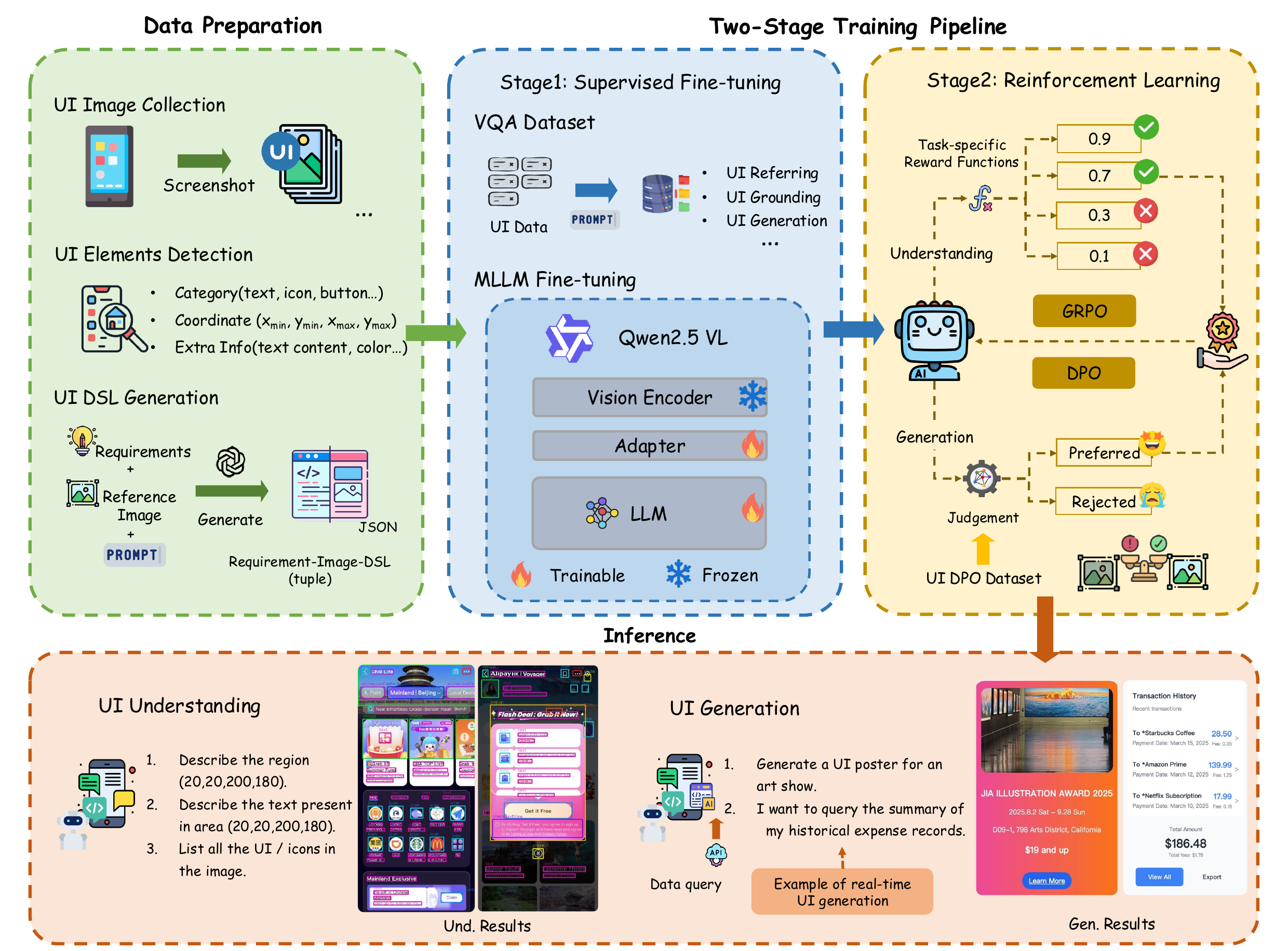} % Reduce the figure size so that it is slightly narrower than the column. Don't use precise values for figure width. This setup will avoid overfull boxes.
\vspace{-0.5em} 
\caption{The workflow for UI-UG includes 1) Data preparation (UI image collection + element detection + DSL generation); 2) Two-stage training: SFT with VQA dataset, then RL optimization using GRPO and DPO for each task. 
The model supports UI understanding tasks (referring and grounding) and enables both offline and real-time UI generation.}
\vspace{-1.1em} 
\label{fig2}
\end{figure*}

With the continued increase in UI resolution and improvement of design complexity, the challenges in UI-related tasks have escalated significantly. As shown in Figure \ref{fig1}, modern UIs now tend to feature abundant commercial content (banner, pop-up windows) and pack UIs in a dense way (smaller arranged icons, nested elements). Moreover, this evolution demands a finer-grained understanding of UI elements, for instance, customized icon recognition requirements for specialized icons (e.g., 'close', 'return', 'more') to support automated agent operations. Consequently, legacy open-source UI datasets such as RICO \cite{rico} and VINS \cite{bunian2021vins} have become increasingly outdated in visual style and insufficient to meet current demands. Compared to UI understanding tasks, generation tasks present a more comprehensive and complex challenge. They require not only image-type perception but also a deeper global understanding of UI hierarchical structures. While most prior work Pix2Code \cite{beltramelli2018pix2code}, Sketch2Code \cite{jain2019sketch2code}, DCGen \cite{dcgen}, DeclarUI \cite{declareUI}, only focused on offline or development-phase generation, the era of multimodal large language models enables real-time, interactive UI generation during conversational workflows.

In this paper, we propose \textbf{UI-UG}, the first unified MLLM for \textbf{UI Understanding and Generation}. Figure \ref{fig2} illustrates the whole pipeline of our work, including data preparation, two-stage model training and model inference.

To address these issues above, we first enhanced our dataset by collecting screenshots of over 30,000 images from modern mobile apps, covering homepages, feeds, marketing modules, etc. We designed our fine-grained UI classification categories, including pop-up windows, functional icons, etc., and added advanced tasks like OCR and color recognition to expand our UI understanding capabilities. Building upon this enhanced understanding capability, we started our core task of UI generation. Since UI generation is about producing renderable code instead of generating images, we designed a frontend-friendly domain-specific language (DSL) in JSON format containing UI types, hierarchical structures, mock data placeholders, and CSS tokens. Our DSL design allows the model to consume data queried from the backend during user interactions, which supports dynamic, real-time, and incremental UI rendering. Next, we used larger MLLMs with customized-designed system prompts to generate requirements and corresponding DSLs from UI images, and then further constructed the UI generation dataset (Requirement-Image-DSL tuples) by validating DSL rendering and selecting high-quality renders.

Subsequently, based on fundamental annotations for UI understanding and generation, we constructed a VQA dataset, which includes referring, grounding, and generating tasks. We started our fine-tuning with Qwen2.5-VL-7B model, while experimenting with data processing and reinforcement learning methods like GRPO \cite{deepseek-r1} and DPO \cite{dpo} to further improve performance. For instance, we enhanced detection accuracy by grounding sorting and alignment techniques, leveraging the regularity and repetitive patterns in UI. And we optimized referring and grounding tasks by GRPO using format, accuracy, and IoU rewards.
% , and box alignment reward. 
Additionally, we improved the visual quality and stability of generated outputs through our well-designed DPO preference dataset.

 We compared our model UI-UG with current general-purpose MLLMs (e.g., GPT-4o, Qwen2.5-VL-72B) and UI-specific models (e.g., OmniParser \cite{omniparser,omniparserv2}, Ferret-UI2 \cite{ferret-ui2}) on understanding tasks, and our model achieved markedly superior performance. Finally, we also observed the enhancements on our generation task through reinforcement learning optimization, outperforming the generation capabilities of leading general-purpose large models. During our experiments, we also observed that combining UI understanding and generation tasks in training yielded better performance than training them separately, which indicates a potential synergistic effect between these two tasks.
% Notably, even when initialized from baseline models without specialized training, our model demonstrated strong capabilities through training on our high-complexity dataset.

In summary, our key contributions are:
\begin{itemize}
\item  We proposed \textbf{UI-UG}, a first unified MLLM for UI understanding and generation, tailored for modern UI tasks.
\item  We broadened the task scope to advance UI structural comprehension, and used more methods (SFT, GRPO, DPO) to improve performance for both understanding and generation tasks.
\item  We developed a practical workflow for real-world applications, including LLM-friendly DSL design, generating, evaluation, and progressive rendering capabilities. 
\item  Our model achieved SOTA performance on modern complex UI understanding tasks, while attaining comparable UI generation quality to larger models at a fraction of their computational cost.
\end{itemize}

\section {Related Work}

\textbf{UI Understanding.} 
Understanding UIs is fundamental for intelligent automation and enhancing human-computer interaction. 
% Research in this area has evolved significantly, driven by advancements in datasets and modeling paradigms.
Early efforts were supported by large-scale datasets like Rico \cite{rico}, ERICA \cite{erica}, while later benchmarks like WebSRC \cite{chen2021websrc} and VINS \cite{bunian2021vins} introduced more complex challenges. Traditional methods understand UI from structural methods that parsed Document Object Models (DOM) or view hierarchies \cite{cai2003vips, song2004learning}.
The advent of computer vision and large multimodal models, models like ScreenParsing \cite{wu2021screenparsing}, Pix2Struct \cite{lee2023pix2struct} and the vision-based OmniParser \cite{omniparser,omniparserv2} demonstrated parsing UIs directly from pixels. For a more granular understanding, specialized models like Ferret-UI \cite{ferret-ui} and Ferret-UI2 \cite{ferret-ui2} unified various UI tasks across platforms.
% The powerful vision capabilities of generalist models like GPT-4o \cite{gpt4o} and the GUI-centric design of CogAgent \cite{hong2024cogagent} have established new state-of-the-art standards. Building on this, 
% There has been an increasing focus on autonomous GUI agents. Models like SeeClick \cite{cheng2024seeclick}, Iris \cite{ge2024iris}, UI-Tars \cite{qin2025ui-tars} specifically target the crucial step of mapping instructions to clickable elements. 
In this work, we adopted MLLMs and built upon similar frameworks established in prior studies, and conducted comparisons with these works.

\textbf{UI Generation.} Automated UI code generation speeds up development by automatically turning visual designs into working code. 
% Driven by the need for faster app creation and making apps that work on different platforms, this technology uses AI to build UIs directly from an image or a demand.
% Current research is advancing along three primary directions: deep learning (DL), computer vision (CV), and multimodal large language models (MLLMs).
DL-based approaches like Pix2code \cite{beltramelli2018pix2code} used CNNs and LSTMs to convert UI images into code, often a Domain-Specific Language (DSL). In parallel, CV-based methods like Sketch2Code \cite{jain2019sketch2code} and REMAUI \cite{nguyen2015remaUI} focused on explicitly identifying UI components. 
% They use techniques like object detection on sketches or OCR on mockups to detect elements and infer their layout structure before generating code.
The latest methods utilize MLLMs to overcome persistent issues like element omission and misarrangement. For instance, DCGen \cite{dcgen} and DeclarUI \cite{declareUI} employ a divide-and-conquer approach, prompting an MLLM to generate code for submodules separately and then assembling the final webpage. Works like Web2Code\cite{yun2024web2code}, LiveCodeBench \cite{jain2024livecodebench} also build benchmarks for evaluating UI generation from LLMs. In our work, we leveraged MLLMs to generate UI DSL code from user requirements and reference images, with quantitative evaluation against benchmark models from prior studies.

\textbf{Reinforcement Learning with MLLMs.}
The fine-tuning of large models has shifted from traditional Reinforcement Learning from Human Feedback (RLHF) \cite{christiano2017rlhf,ziegler2019rlhf2} towards more direct approaches like Direct Preference Optimization (DPO) \cite{dpo}, eliminating the complex reward-modeling step.
Group Relative Policy Optimization (GRPO) \cite{deepseek-r1} advances AI training by using a more efficient group-wise optimization to find the best-of-N response, significantly boosting the performance and alignment of LLMs. These reinforcement learning techniques also extend powerfully to Multimodal Large Language Models (MLLMs). Key developments include Visual-RFT \cite{liu2025visual-rft}, applying visual reinforcement fine-tuning for vision-language tasks; R1-VL \cite{zhang2025r1-vl} and VLM-R1 \cite{shen2025vlm-r1}, utilizing step-wise GRPO for stable and generalizable multimodal reasoning. 
% Reinforcement learning also enhances GUI agents, as seen in UI-R1\cite{lu2025ui-r1} for efficient action prediction. 
In this work, we implemented the aforementioned reinforcement learning methods (GRPO/DPO) and adapted them to multimodal UI task frameworks.
% These works collectively advance MLLM capabilities in reasoning, safety, and efficiency using refined RL paradigms.

\section {Data Preparation}
For UI understanding and generation tasks, we constructed two datasets: UI Elements Dataset and UI DSL Dataset.

\subsection{UI Elements Dataset} 

\subsubsection{UI Screenshots Collection.}First, we collected screenshots randomly from widely-used, popular apps. Notably, to overcome the limitation of standard screenshots often truncating UI contents, we developed engineering solutions to generate full-length, scrollable screenshots. Subsequently, we encoded the images using the CLIP \cite{clip} model and performed deduplication based on image embedding similarity. Ultimately, our dataset contains over 30,000 images. All images have resolutions within the range of [380*720, 1550*3800], with approximately 76\% being standard smartphone-sized images and 24\% being full-length scrollable screenshots.

\subsubsection{UI Element Classes.}Building upon basic UI element categories (text, image, icon, button), we introduced higher-level semantic UI classes that frequently appear in our collected screenshots, such as pop-up and progress bar. Additionally, to better address agent-oriented tasks, we incorporated a deeper semantic understanding of screen contents, including the identification of interactive elements such as functional icons and checkboxes.

\subsubsection{UI Element Annotation.}To accelerate the annotation process, we first employed crowdsourced manual labeling on approximately 3,000 samples (10\% of the dataset). Using this annotated data, we trained YOLOv8 \cite{yolov8_ultralytics} grounding models to generate bounding box annotations automatically for the remaining data. Afterwards, we further predicted textual content using OCR\cite{du2021ppocrv2} methods and text color via color clustering for text elements, aiming to enrich the dataset with fine-grained content and stylistic information. Finally, human annotators refined these automated annotations to construct the high-quality final UI dataset.

\subsection{UI DSL Dataset} 

\begin{figure}[t]
\centering
\includegraphics[width=0.7\columnwidth]{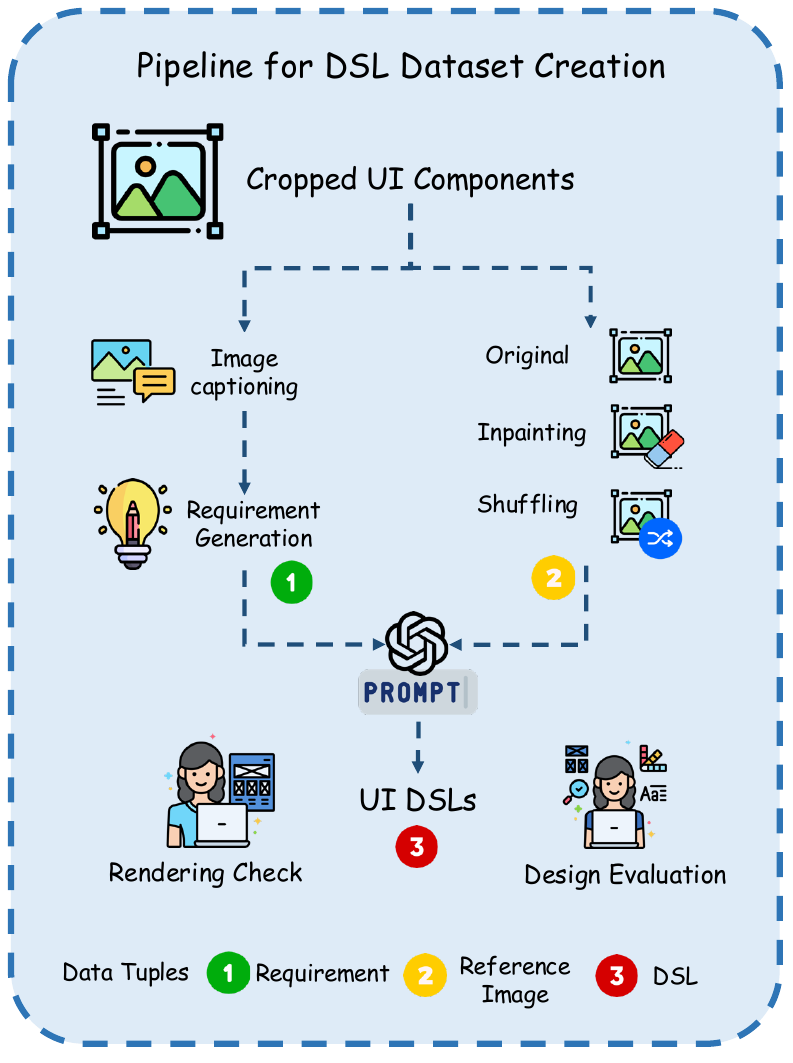} 
\vspace{-0.5em} 
\caption{Pipeline for UI DSL Dataset creation. The final data tuples consist of requirements, reference images, and corresponding UI DSLs.
}
\vspace{-1.3em} 
\label{fig3}
\end{figure}

\subsubsection{UI Element Cropping and Description.}In practical industrial scenarios for UI generation tasks, engineers typically focus on generating individual UI components or their combinations rather than reconstructing entire pages. To align with this workflow, we cropped the original screenshots using a UI component detector. Subsequently, we utilized Qwen2.5-VL-72B to generate captions for the cropped UI components to further simulate user requirements.

\subsubsection{DSL Design.}Our UI DSL utilizes JSON as its core structure. For detailed UI style descriptions, it incorporates Tailwind CSS-like tokens such as color, font and alignment. Besides, a key feature of our DSL is mock fields, which allow engineers to dynamically replace text, data or images at runtime via dynamic inputs. This design also leverages the prevalence of JSON partial parsers, enabling streaming-based progressive rendering even before the full UI is completely generated.

\subsubsection{DSL Generation.}Considering the cost, we utilized the open-source models Qwen2.5-VL-72B to generate DSL specifications corresponding to each UI component. For each input, we treated the generation requirement as a query and an optional UI image as a reference for color schemes, styles, and layout. In the system prompt, beyond basic task instructions, we explicitly defined our DSL structure, demonstration examples to guide the MLLM to strictly adhere to the UI DSL specifications.

\subsubsection{Style Mixing.}To enhance the model's style transfer capabilities, we randomly applied inpainting or non-inpainting operations to reference images, and performed random combinations of reference images and user requirements. This approach not only improves style adaptability but also prevents textual content in reference images from dominating generation results during subsequent model training.

\subsubsection{DSL Validation.}We employed rendering checks (using a UI DSL renderer) and designer evaluations to validate and filter suitable generated DSL outputs. Finally, this process generated a dataset of tuples containing UI requirements, reference images, and corresponding UI DSL specifications as shown in Figure \ref{fig3}.

\section {Method}
\subsection{Supervised Finet-tuning (SFT)}
\subsubsection{VQA Dataset.}
We provided 10 prompt templates per task to generate Visual Question Answering (VQA) data for referring, grounding, and generating tasks. For consistency and easy parsing, all final dataset results are in JSON format. All images are smart resized following Qwen2.5-VL's image processing method, scaling image dimensions to the nearest multiple of 28 while adhering to min pixels with 64*28*28 and max pixels with 1280*28*28 constraints. To enhance the model's spatial perception, all coordinates were wrapped with special tokens, specifically $<box\_start>(x_{min}, y_{min}),(x_{max}, y_{max})<box\_end>$.

For grounding tasks, considering the structural regularity and alignment in UI elements (e.g., icons or components arranged in grid-like layouts), we first grouped bounding boxes by category and sorted them by category frequency. Subsequently, we ordered them by left-top coordinates. According to ROPE \cite{su2024rope} position embedding, coordinate outputs for adjacent regions tend to be encoded with rotational transformations that preserve their relative spatial order. This ensures adjacent boxes exhibit greater proximity in category or spatial structure, enabling the model to better capture attention on relevant elements and reduce missed or false detections.
For generating tasks, we additionally employed LLMs to rewrite requirements, condensing them into concise one-sentence or multi-sentence descriptions. This improves adaptability for ambiguous user requirements. During dataset construction, we also generated both image-referenced and non-image-referenced datasets.

\subsubsection{Model Fine-tuning.} 
We post-train our model \textbf{UI-UG-7B-SFT} by fine-tuning from Qwen2.5-VL-7B model on our VQA dataset for understanding and generation tasks. During training, we froze the ViT \cite{dosovitskiy2020vit} module and trained the Large Language Model (LLM) and Vision-Language Adapter (VL Adapter) modules, setting the max learning rate for 1e-5, and max-tokens for 8192. We also employed a standard warmup-stable-decay learning rate schedule to ensure stable training and better convergence. Through fine-tuning, we sampled approximately 180,000 training VQA data and 0.8 billion tokens in total. The data ratio was approximately 4:3:2, and the trained tokens ratio was 2:6:2 for each task (referring:grounding:generating). The training process for 3 epochs took about 2 days on 8 NVIDIA A100 GPUs with a batch size of 2 per GPU (total global batch size is 16).

\subsection {Reinforcement Learning (GRPO, DPO)}
Although the model has achieved decent results on various tasks during the SFT phase, the success rate and quality have not yet reached a particularly high standard. For example, the model fails on rare UI components in referring tasks and lacks detail and precise alignment in grounding tasks. There is also a need to further enhance the visual quality of the generations. Given that the generated outputs may be lengthy, the stability of format following also requires improvement. To address these issues, we designed a series of verifiable and task-specific rewards and trained our model using the GRPO method for understanding tasks. And we constructed a preference dataset and trained our model using the DPO method for generation tasks.

\subsubsection{Referring.}
We observed that significant data imbalance adversely affected the model's classification performance during the SFT phase, where text, icons, and images account for over 80\%. Thus, we identified 8 challenging component categories (SelectabletextButton, closeButton, popup, checkedText, checkTextView, etc.) to further strengthen. Additionally, we defined hard samples by manually selecting difficult samples and running inference five times with the SFT model; if it is incorrect even once, it is considered a hard sample.

 Next, we defined the rewards, primarily focusing on classification accuracy for these challenging component categories and adherence to JSON format. 
% Notably, the improvement in classification performance for the Referring task also led to enhancements in downstream tasks, such as OCR results.
 The final GRPO reward for the referring task is defined as follows: \\
\begin{equation}
  R_{\text{ref}} = R_{\text{acc}} + R_{\text{format}} \label{eq:energy}
\end{equation}

 \subsubsection{Grounding.} 
 We used dual Intersection over Union (dual IOU) \cite{wang2025traceableground} to measure the quality of predicted boxes against ground-truths. Specifically, the dual IOU function consists of two parts (recall IOU reward $R_{\text{IOU}}^{R}$ and precision IOU reward $R_{\text{IOU}}^{P}$).

 % Specifically, after each prediction, we perform bipartite graph matching with the ground truth and use dual IOU \cite{wang2025traceableground} to measure the quality of predicted boxes against ground-truths (dual IoU is an average of a recall term and a precision term). TODO Align. The GRPO reward function for grounding is defined as: \\
 % \begin{equation}
 %  R_{ground} = 1/2*(R_{IOU}^R + R_{IOU}^P) + R_{align} + R_{format}   \label{eq:energy}
% \end{equation}
for $N$ ground-truth bounding boxes $\left\{\boldsymbol{b}_i\right\}_{i=1}^N$ and
$M$ predicted bounding boxes $\left\{\hat{\boldsymbol{b}}_j\right\}_{j=1}^M$, The recall IOU reward is determined as follows:
 \begin{equation}R_{\text{IOU}}^{R} =\frac{1}{N} \sum_{i=1}^N \operatorname{IoU}\left(\hat{\boldsymbol{b}}_j, \boldsymbol{b}_i\right)\label{eq:energy}\end{equation}

which means each ground-truth box $b_i$ is greedily matched with the predicted bounding box $\hat{\boldsymbol{b}}_j$ yielding the highest  (IoU). Similarly, we also define a precision IoU reward, $R_{\text{IOU}}^{R}$, by greedily matching each predicted box $\hat{\boldsymbol{b}}_j$ with the ground-truth box that yields the highest IoU:
 \begin{equation}R_{\text{IOU}}^P = \frac{1}{M} \sum_{j=1}^M \operatorname{IoU}\left(\boldsymbol{b}_i, \hat{\boldsymbol{b}}_j\right)\label{eq:energy}\end{equation}
 
 According to the practice of \cite{luo2025guir1}, we didn't train the chain-of-thought(CoT) for the grounding task to get better performance. As JSON format reward is also concerned, the final GRPO reward for the grounding task is defined as follows: 
 \begin{equation}R_{\text{ground}} =R_{\text{IOU}}^{R} + R_{\text{IOU}}^P + R_{\text{format}} \label{eq:energy}\end{equation}

 \subsubsection{Generation.}
  We referred Web2Code \cite{yun2024web2code}, for images rendered from UI generations, employed Qwen2.5-VL-72B to query scores across four categories scores as follows: 1) Visual Structure \& Alignment, 2) Color \& Aesthetic Design, 3) Textual \& Content Consistency, 4)Interface \& Interactivity.
% \begin{enumerate}[itemsep=1pt, topsep=1pt, parsep=0pt, partopsep=0pt, leftmargin=0.85cm]
% \item Visual Structure \& Alignment 
% \item Color \& Aesthetic Design
% \item Textual \& Content Consistency
% \item Interface \& Interactivity
% \end{enumerate}
  The scores for each category range from 0 to 10. 
  We then constructed a DPO dataset totaling  8,000 samples, where positive samples primarily consist of high-scoring rendered data, and negative samples include low-scoring rendered data as well as data that failed rendering due to incorrect JSON format. We observe that using only the DPO component may lead to a decrease in the probability of positive samples. Inspired by work \cite{wang2024enhancing}, we additionally introduced SFT loss, which helps the model learn how to generate high-quality responses.

  The DPO loss function is defined as follows:

\begin{equation}
\mathcal{L}_{\text{total}} = \mathcal{L}_{\text{DPO}} + \lambda \mathcal{L}_{\text{SFT}}
\end{equation}

\begin{table*}[t]
\centering
\begin{tabular}{l c c c c c c c c}
\toprule
 \textbf{Model}  & \multicolumn{4}{c}{\textbf{\textit{Referring}}} &
\multicolumn{4}{c}{\textbf{\textit{Grounding}}} \\
\cmidrule(lr){2-5} \cmidrule(lr){6-9}
 & Format & Category & OCR & TextColor & Format & mAP & AP50 & AP75 \\
 % Model&Format& Category& OCR& TextColor& mAP& AP50& AP75& mIOU50&mIOU75\\

\midrule
 GPT-4o& 1& 0.273& 0.193& 0.350& 0.932 & 0.018& 0.029& 0.008\\
 Claude 3.7 Sonnet& 0.953& 0.680& 0.400& 0.462& 0.939 & 0.010& 0.020& 0.001\\
 Gemini 2.5 Pro& 1& 0.400& 0.157& 0.514& 0.959 & 0.013& 0.021& 0.004\\
Qwen2.5-VL-7B&0.419& 0.527& 0.586& 0.649& 0.923& 0.092& 0.120& 0.064\\
 Qwen2.5-VL-32B&0.317& 0.754& 0.841& 0.447& 0.918& 0.183& 0.258& 0.107\\
 Qwen2.5-VL-72B& 0.575& 0.744& 0.826& 0.686& 0.918& 0.266& 0.345& 0.187\\
 \midrule
 Ferret-UI2 (Gemma-2B)& 1& 0.789& -& -& 0.715 & 0.070& 0.108& 0.032\\
 Ferret-UI2 (Llama3-8B)& 1& 0.758& -& -& 0.855 & 0.083& 0.119& 0.046\\
 OmniParser V2& -& -& -& -& -& 0.290& 0.382& 0.198\\

 \midrule
 UI-UG-7B-SFT ($\text{U}_\text{{Only}}$& 1& 0.968 & 0.850 & 0.932 & 1& 0.527 & 0.547& 0.507\\
 UI-UG-7B-SFT& 1& 0.972& \textbf{0.859}& \textbf{0.937}& 1& 0.544& 0.567& 0.527\\
 UI-UG-7B& 1& \textbf{0.974}& 0.854 & 0.933 & 1 & \textbf{0.559} &\textbf{0.578} & \textbf{0.540} \\

\bottomrule
\end{tabular}
\vspace{-0.3em} 
\caption{Performance comparison of different models on referring and grounding tasks.}
\vspace{-1em} 
\label{tab:grounding}
\end{table*}

DPO Loss Component:
$$
     \mathcal{L}_{\text{DPO}} = -\mathbb{E}_{(q, y^w, y^l)\sim D} \log 
$$

\begin{equation}
\sigma\left[ \beta \log \frac{\pi_\theta(y^w|q)}{\pi_{ref}(y^w|q)} - \beta \log \frac{\pi_\theta(y^l|q)}{\pi_{ref}(y^l|q)}\right]
\end{equation}

SFT Loss Component:
\begin{equation}
\mathcal{L}_{\text{SFT}}(\theta) = -\mathbb{E}_{(x,y) \sim \mathcal{D}_{\text{SFT}}} \big[ \log \pi_\theta(y \mid x) \big]
\end{equation}

where $ \pi_\theta$ and $\pi_{ref}$ denote the learnable and reference model,
 $y^w, y^l\sim\pi_{\theta}(\cdot|q)$ represent winning (preferred) and lossing (rejected) response for query $q$, $\sigma$ is the sigmoid function. $\lambda$ controls balance for SFT loss and DPO loss and $\beta$ controls KL divergence penalty.

%   \begin{equation}
% \mathcal{L}_{\text{DPO}}(\theta) = -\mathbb{E}_{(x,y_w,y_l) \sim \mathcal{D}} \left[ 
% \log \sigma \left( 
% \beta \underbrace{\log \frac{\pi_\theta(y_w \mid x)}{\pi_{\text{ref}}(y_w \mid x)}}_{\text{winning log-ratio}} 
% - \beta \underbrace{\log \frac{\pi_\theta(y_l \mid x)}{\pi_{\text{ref}}(y_l \mid x)}}_{\text{losing log-ratio}} 
% \right) 
% \right]
% \end{equation}
% Where:
% \begin{itemize}
%     \item $\theta$: Parameters of the policy model being optimized
%     \item $x$: Input prompt/context
%     \item $y_w$: Preferred (winning) response
%     \item $y_l$: Rejected (losing) response
%     \item $\mathcal{D}$: Dataset of preference triplets $(x, y_w, y_l)$
%     \item $\pi_{\theta}$: Current learnable policy (language model)
%     \item $\pi_{\text{ref}}$: Reference policy (usually initial SFT model), kept frozen
%     \item $\beta$: Temperature parameter controlling deviation from reference policy
%     \item $\sigma$: Sigmoid function $\sigma(x) = (1 + e^{-x})^{-1}$
%     \item $\mathbb{E}$: Expectation over dataset samples
% \end{itemize}
  
  For the entire reinforcement learning stage, we sequentially applied DPO for generating, GRPO for referring, and GRPO for grounding to generate our final model \textbf{UI-UG-7B}. The learning rate is set to 3e-6,  $\beta$ is set to 0.05, and $\lambda$ is set to 1.0. For GRPO training, we set the learning rate to 1e-6 and the global batch size to 64. We adopt GRPO for 2 epochs and DPO for 3epochs on the same GPUs as SFT phase. The whole process took about 6 hours. 
  % considering that GPRO has a milder impact on the model's parameters.

\section{Experiment}
We conducted comprehensive comparisons with existing models on the UI Elements Dataset across two core tasks: UI understanding and UI generation. Our evaluation encompasses both general-purpose MLLMs (including closed-source models like GPT-4o, Claude 3.7 Sonnet, and Gemini 2.5 Pro, as well as open-source alternatives from the Qwen2.5-VL series) and the latest well-known UI domain-
specific models (Ferret-UI2 and OmniParser V2). 
% Additional results and visual comparisons are presented in the supplementary materials.

% To validate our design choices, we perform ablation studies examining data preparation strategies, supervised fine-tuning optimizations, and the impact of reinforcement learning.
\subsection{UI Understanding Evaluation}
\subsubsection{Evaluation Metrics.}
For UI understanding evaluation, we utilized our modern complex UI dataset described before, focusing on two primary tasks. The referring task requires describing element attributes given ground-truth bounding boxes, evaluated through three metrics: 1) category recognition accuracy for element classification, 2) OCR accuracy computed as $1 - \frac{N_{diff}}{N_{total}}$ where $N_{N_{diff}}$ denotes differing characters, 3) color recognition accuracy measured by $\sum\left[(\frac{r_{pred}-r_{gt}}{255})^2 + (\frac{g_{pred}-g_{gt}}{255})^2 +(\frac{b_{pred}-b_{gt}}{255})^2\right]$. 

We evaluated UI element localization performance for grounding tasks using standard object detection metrics: mean Average Precision (mAP) and Average Precision at IoU thresholds of 0.5 and 0.75 (AP50, AP75). 

We also evaluated the instruction following capabilities by JSON format accuracy for both tasks.

\subsubsection{Implementation Details.}
All general-purpose MLLMs received prompt-based instructions specifying UI categories and output formats with inference temperature 0. We implemented model-specific coordinate normalization: 0-1 range normalization for GPT-4o and Claude 3.7 Sonnet versus 1000*scaling for Gemini 2.5 Pro. For domain-specific models, we developed categorical mappings to align disparate UI taxonomies. Special handling includes excluding Ferret-UI2 from OCR/color tasks due to its lack of multilingual support and limiting OmniParser V2 evaluation to grounding tasks as its traditional vision model architecture.

\subsubsection{Results.}
As summarized in Table \ref{tab:grounding}, our UI-UG-7B model achieved SOTA performance on both referring tasks (Format:1, Category:0.974, OCR:0.854, TextColor:0.933) and grounding tasks (Format:1, mAP:0.559, AP50:0.578, AP75:0.54). Closed-source MLLMs (GPT-4o, Claude 3.7 Sonnet, Gemini 2.5 Pro) performed poorly on referring tasks and exhibited fundamental limitations in spatial understanding and perception for regional information. 
They also demonstrated significant detection misses and coordinate errors in grounding tasks, resulting in predicted bounding boxes deviating substantially from ground-truth. Further granular analysis revealed that these general-purpose models lack recognition and classification capabilities for UI elements, particularly rare ones. Even their performance on the most common elements like text (OCR Grounding) still remained low, aligning with recent findings by other researchers \cite{li2025towardstext}.
In contrast, the open-source Qwen2.5-VL (7B, 32B, 72B) model, which is the foundation for our UI-UG model, possesses stronger capabilities in both referring and grounding tasks. However, it struggles significantly with format accuracy, especially in referring tasks(under 0.6).

Among domain-specific models, Ferret-UI2 (Gemma-2B, Llama3-8B) showed marginal superiority over the best general-purpose MLLM in UI classification tasks. Nevertheless, its referring and grounding performance lagged far behind our model. We attributed this gap partly to its reliance on older datasets (e.g., RICO). OmniParser V2 achieved improved grounding performance compared to Ferret-UI2, yet still remained inferior to our approach (mAP:0.29 vs 0.559).

We also observed that our final model UI-UG, which is reinforcement learning enhanced, demonstrates a more significant improvement compared to the SFT model, especially pronounced for the grounding tasks.

\begin{figure}[t]
\centering
\includegraphics[width=0.97\columnwidth]{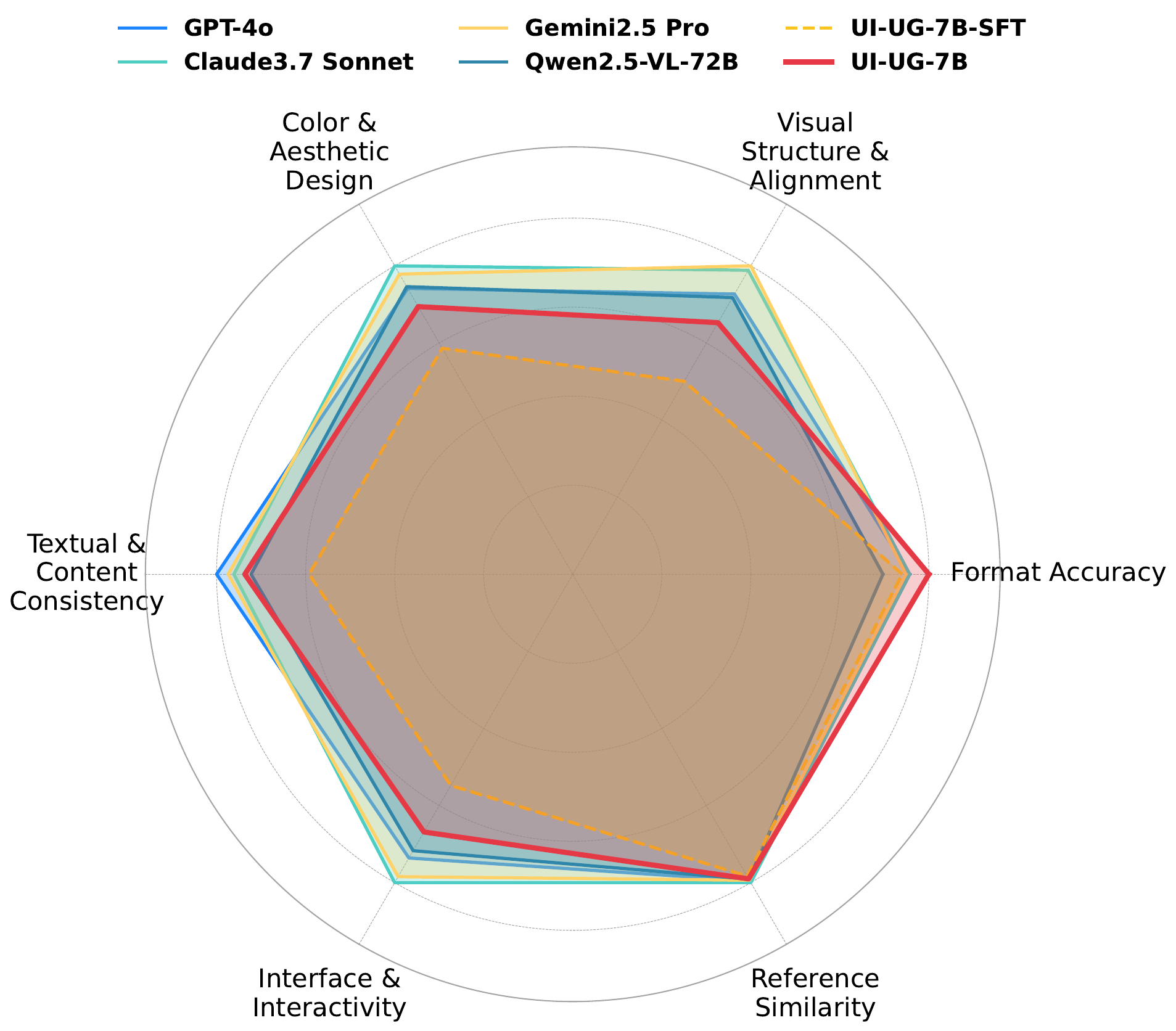} % Reduce the figure size so that it is slightly narrower than the column. Don't use precise values for figure width. This setup will avoid overfull boxes.
\vspace{-0.5em} 
\caption{Generation scores for different models, with each dimension normalized. Our model UI-UG achieved quality enhancements through reinforcement learning and now approaches the level of current powerful larger models.}
\vspace{-1.5em} 
\label{fig4}
\end{figure}

\subsection{UI Generation Evaluation}
\subsubsection{Evaluation Protocol.}
We evaluated UI generation in six key categories (each score rated 0-10, totaling 60) through the following: 1) JSON format compliance rate measuring DSL syntax, 2) CLIP-based visual similarity between generated and reference UIs, and 3) MLLM-based scoring in four categories following Web2Code. 

% Visual Structure \& Alignment, Color \& Aesthetic Design, Textual \& Content Consistency and Interface \& Interactivity .

\subsubsection{Performance Analysis.}
Figure \ref{fig4} presents the radar chart of scores from various MLLMs. Due to training on the UI DSL Dataset, our model achieved the highest score in generative format stability. Following DPO training, our model showed a nearly 14.5\% improvement in MLLM scores across four dimensions (from 36.7 to 42.02), demonstrating the effectiveness of our preference learning via DPO. Although its generative quality did not surpass that of closed-source models (Claude 3.7 Sonnet: 45.08, GPT-4o: 43.76, Gemini 2.5 Pro: 44.80), it performed on par with the model that produced training data (Qwen2.5-VL-72B: 42.15). 

Generating speed was also measured. By fine-tuning a smaller model (7B), we reduced system prompt tokens and achieved lower first token latency with faster model inference. In our tests, our model deployed on two NVIDIA L20 GPUs completed a full output on average of 5.2s, while other models required 20-30s. The GPTQ / AWQ-INT4 quantized version can further reduce inference time to under 2s, enabling real-time generation during user conversations.

\subsection{Ablation Study}
Table \ref{tab:ablation} shows the result of our additional ablation study, where theudy, where the, where the scores for each task are aggregated. First, we experimented with trainable parameters for SFT, including: only LLM, Adapter+LLM, and ViT+Adapter+LLM. The Adapter+LLM strategy achieved the best-balanced performance and is adopted in the next experiments.

Next, we compared models trained exclusively on UI understanding data versus UI generation data. Incorporating UI understanding data enhanced both the quality and score of UI generation, while the model trained solely on understanding data showed no significant difference in understanding tasks. This confirms that jointly training understanding and generation tasks improves model accuracy more effectively through their synergistic knowledge transfer.
%  ( GenScore +0.5\%)
% (RefScore -1.7\%,  mAP +3.2\%)

In reinforcement learning experiments, both GRPO and DPO demonstrated significantly improved performance in their respective domains (RefScore +1.2\%, mAP +4.6\%,  GenScore +15.9\%) compared to the SFT model.

Our final model, which combines SFT, GRPO, and DPO, achieved the most balanced excellent performance for all tasks (RefScore: 3.761, mAP: 0.559,  GenScore: 42.02).

% Notably, the improvement in classification performance for the Referring task also led to enhancements in downstream tasks, such as OCR results.

\begin{table}[t]
\centering
\hspace*{-0.3cm} 
% \begin{tabular}{l c c c }
\begin{tabular}{@{}l c@{\hspace{1pt}}c@{\hspace{1pt}}c@{}}

\toprule
Model & \textbf{\textit{Referring}}& \textbf{\textit{Grounding}}& \textbf{\textit{Generating}}\\
 \cmidrule(lr){2-2} \cmidrule(lr){3-3} \cmidrule(lr){4-4}
 & RefScore& mAP& GenScore\\

 \midrule
SFT$_\text{{L}}$& 3.685& 0.489& 36.65\\
 SFT$_\text{{A+L}}$& 3.731& \textbf{0.544} & 36.70\\
 SFT$_\text{{V+A+L}}$& \textbf{3.740} & 0.451& \textbf{36.94}\\
 \midrule
  SFT$_\text{{A+L}}$ (\text{U}$_\text{{Only}}$)& 3.792& 0.527&-\\
SFT$_\text{{A+L}}$ (\text{G}$_\text{{Only}}$)& -& -& 36.52\\
 \midrule
 SFT$_\text{{A+L}}$\text{+GRPO}$_\text{U}$& \textbf{3.777}& \textbf{0.569}& 37.34\\
 SFT$_\text{{A+L}}$\text{+DPO}$_\text{G}$& 3.764& 0.535 &\textbf{42.53}\\
  \midrule
 SFT$_\text{{A+L}}$\text{+Both(Final)}  & 3.761& 0.559 & 42.02\\
\bottomrule
\end{tabular}
\vspace{-0.3em} 
\caption{Ablation study on training strategy. Final model UI-UG used SFT(A+L) + Both (GRPO for Understanding, DPO for Generation), and achieved the most balanced excellent performance. (L: LLM, A: Adapter, V: VIT)}
\vspace{-1.4em} 
\label{tab:ablation}
\end{table}

\subsection{Extended Zero-shot Evaluation}
Furthermore, we conducted two additional zero-shot experiments to validate model generalizability. First, we randomly sampled 130 RICO images to build a streamlined UI understanding benchmark. Our model maintains superiority with 82\% category classification accuracy versus Ferret-UI2 (Llama3-8B)'s 70\%, with 0.15 detection mAP versus Ferret-UI2's 0.08 and OmniParser V2's 0.11. Visualization of annotations further verifies that the performance decline in this benchmark stems from RICO's erroneous labeling rather than model overfitting. 

Additionally, we evaluated captioned grounding task on ScreenSpot \cite{cheng2024seeclick} benchmark, despite having no such training data. However, this UI-related task showed a 24.8\% decrease (from 84.7 to 63.7) in localization metrics compared to our base model (Qwen2.5-VL-7B). We attribute this to the inherent opposition between captioning (describing UI) and grounding (generating UI from descriptions), which prevents knowledge sharing, unlike the complementary tasks of UI understanding and generation. More future work and investigation will focus on this issue.

\section{Conclusion}
We propose UI-UG, a unified MLLM that achieves state-of-the-art (SOTA) performance on modern complex UI understanding tasks and matches the UI generation performance of larger MLLMs at significantly lower computational cost.

As well, we are planning to open-source the model, evaluation benchmarks, and UI rendering toolkits in the near future to support community development. 

\bibliography{aaai2026}

% \newpage

\section {Appendix}
\subsection{1. Details for Data Preparation}

\subsubsection{Definition of UI Categories.}

\begin{figure}[t]
\centering
\includegraphics[width=0.9\columnwidth]{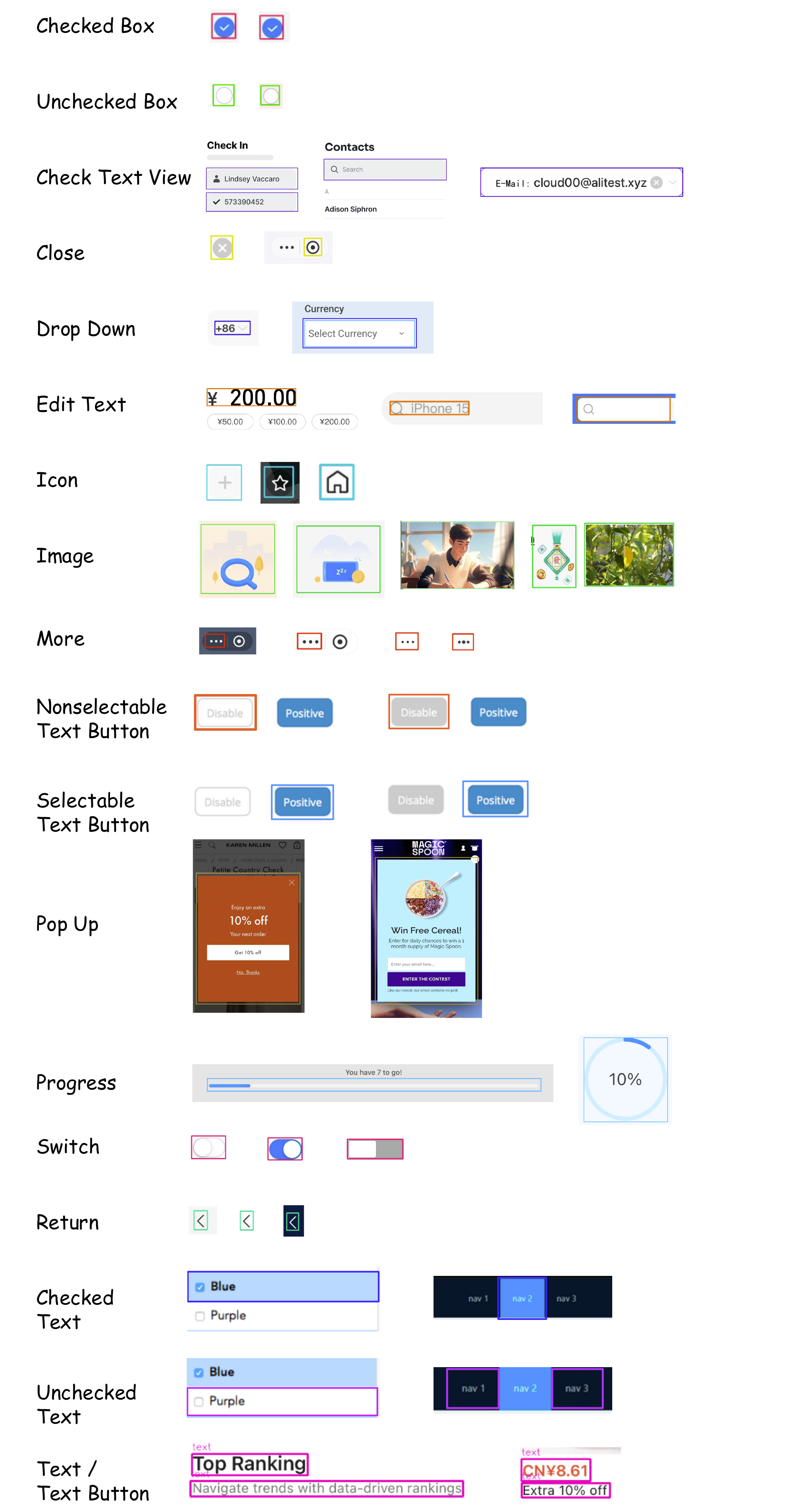} % Reduce the figure size so that it is slightly narrower than the column. Don't use precise values for figure width.This setup will avoid overfull boxes.
\caption{Display of all our UI categories.}
\label{figcat}
\end{figure}

Figure \ref{figcat} presents all UI element categories we designed along with their examples, including a total of 19 distinct types ('Checked Box', 'Unchecked Box', 'Check Text View', 'Close', 'Drop Down', 'Edit Text', 'Icon', 'Image', 'More', 'Nonselectable Text Button', 'Selectable Text Button', 'Pop Up', 'Progress', 'Switch', 'Return', 'Checked Text', 'Unchecked Text', 'Text', 'Text Button').

\subsubsection{Prompts for UI DSL generation.} 
\begin{quote}
\begin{small}
You are an experienced UI designer and a skilled front-end developer. Based on the user's input, return a UI DSL that conforms to the following definition:

UI DSL Definition:

UI DSL is a domain-specific language in JSON format that describes UI content features and styling features.
The basic node attributes of UI DSL include: "type", "name", "className", "params", and "children".
type: Indicates the node type, with optional values being "Component" or "Tag".
name:
If type is "Component", this attribute is the name of a component from the "available components" list.
If type is "Tag", this attribute is the name of an HTML tag.
className: Tailwind CSS class names. Use relative layouts (avoid fixed). Width/height can be absolute (e.g., w-[100px]) or relative (e.g., w-full, w-screen, w-[100\%]).
params: Represents the content features of the node.
If type is "Component", the "value" attribute in params corresponds to the component's property values.
If type is "Tag", params contains the attributes and values of the HTML tag.
The params node must include a "bindType" attribute with possible values "Static" or "Data":
"Static": The attribute is static.
"Data": The attribute is dynamic, referencing a variable from the "real data" (e.g., promotions[0].title).
children: A list of child nodes.

Example Node Data:

json

\{"type": "Tag", "name": "a", "className": "color-gray-500", "params":\{"textContent":\{"bindType": "Static", "value": "Test Link"\}\}\}  

\{"type": "Tag", "name": "a", "className": "color-gray-500", "params":\{"textContent":\{"bindType": "Data", "bindField": "promotions[0].title"\}\}\}  

Design Style Reference:

Button Styles:
Base structure: inline-flex items-center justify-center font-medium transition-colors focus:outline-none focus:ring-2 focus:ring-offset-2 disabled:opacity-50 disabled:pointer-events-none
Primary button: bg-blue-600 text-white hover:bg-blue-700 active:bg-blue-800 rounded-3xl py-2 px-4 border-0
Secondary button: bg-white text-gray-700 hover:bg-gray-50 active:bg-gray-100 rounded-3xl py-2 px-4 border border-gray-300
Output Requirements:

If the "available components" list is empty, only use "Tag" nodes (no "Component" nodes). If components are available, prioritize using them where appropriate.
Use the provided "real data" to populate the UI DSL content dynamically—reference variables instead of hardcoding values. Ensure all dynamic "bindField" paths correctly match the data structure.
If the user provides a UI screenshot, match the generated layout to the element sizes and visual style in the image. Otherwise, default to mobile-friendly UI card dimensions.
The output must have a single root node. Do not include comments. Ensure the result is valid JSON. Avoid direct image URLs—describe image content in detail instead.
Omit the "params" or "children" attributes if they are empty.
To ensure aesthetic appeal, you may adjust content (add/modify/delete), but avoid emojis. Use tags and gradient backgrounds to enhance design quality.
The UI DSL must adhere to:
·Clear content hierarchy
·Subtle rounded corners and shadows
·Consistent spacing
·Clear interactive feedback
·Flat yet slightly dimensional design
·Strong usability and readability
·Modular component design

User Input:

Demand Description: \{demand\} \\
Real Data: \{mock\_json\}
\end{small}
\end{quote}

\begin{figure}[t]
\centering
\includegraphics[width=0.6\columnwidth]{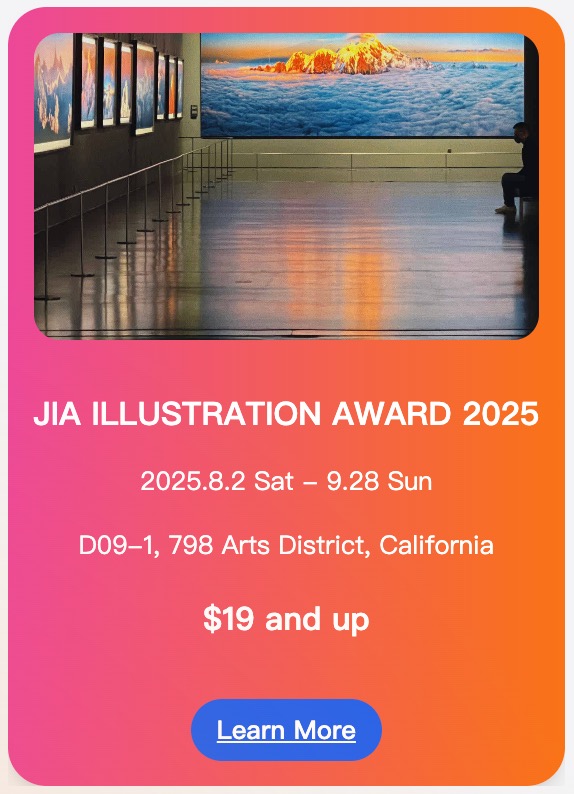} % Reduce the figure size so that it is slightly narrower than the column. Don't use precise values for figure width.This setup will avoid overfull boxes.
\caption{Rendered UI Image by our DSL specification.}
\label{figdsl}
\end{figure}

\subsubsection{Outputs for UI DSL generation.} 

\begin{quote}
\begin{small}
\{"type": "Tag", "name": "div", "className": "flex flex-col items-center justify-center bg-gradient-to-r from-pink-500 to-orange-500 rounded-3xl shadow-lg w-full max-w-sm p-4", "children": [\{"type": "Tag", "name": "img", "className": "w-full h-48 object-cover rounded-2xl mb-4", "params": \{"src": \{"bindType": "Static", "value": "mockImageUrl"\}\}\}, \{"type": "Tag", "name": "div", "className": "text-center", "children": [\{"type": "Tag", "name": "h2", "className": "text-xl font-bold text-white mb-2", "params": \{"textContent": \{"bindType": "Data", "bindField": "title"\}\}\}, \{"type": "Tag", "name": "p", "className": "text-base text-white mb-2", "params": \{"textContent": \{"bindType": "Data", "bindField": "eventDescription"\}\}\}, \{"type": "Tag", "name": "p", "className": "text-base text-white mb-2", "params": \{"textContent": \{"bindType": "Data", "bindField": "dateRange"\}\}\}, \{"type": "Tag", "name": "p", "className": "text-base text-white mb-2", "params": \{"textContent": \{"bindType": "Data", "bindField": "location"\}\}\}, \{"type": "Tag", "name": "p", "className": "text-xl font-bold text-white", "params": \{"textContent": \{"bindType": "Data", "bindField": "price"\}\}\}]\}, \{"type": "Tag", "name": "a", "className": "mt-4 inline-flex items-center justify-center font-medium transition-colors focus: outline-none focus: ring-2 focus: ring-offset-2 disabled: opacity-50 disabled: pointer-events-none bg-blue-600 text-white hover: bg-blue-700 active: bg-blue-800 rounded-3xl py-2 px-4 border-0", "params": \{"href": \{"bindType": "Static", "value": "\#"\}, "textContent": \{"bindType": "Static", "value": "Learn More"\}\}\}]\}
\end{small}
\end{quote}

In addition, Figure \ref{figdsl} presents the rendered UI image by this DSL output.

\subsubsection{2. Prompt Templates} 

\subsubsection{Prompts for Creating VQA Dataset.}

\begin{table*}[t]
\centering
\begin{tabular}{l c c c c c c c c c}
\toprule
 \textbf{Model}  & \multicolumn{4}{c}{\textbf{\textit{Referring}}} &
\multicolumn{4}{c}{\textbf{\textit{Grounding}}} \\
\cmidrule(lr){2-5} \cmidrule(lr){6-9}
 & Format & Category & OCR & TextColor & Format & mAP & AP50 & AP75 \\
 % Model&Format& Category& OCR& TextColor& mAP& AP50& AP75& mIOU50\\

\midrule
 GPT-4o& 1& 0.273& 0.193& 0.350& 0.932 & 0.018& 0.029& 0.008\\
 Claude 3.7 Sonnet& 0.953& 0.680& 0.400& 0.462& 0.939 & 0.010& 0.020& 0.001\\
 Gemini 2.5 Pro& 1& 0.400& 0.157& 0.514& 0.959 & 0.013& 0.021& 0.004\\
Qwen2.5-VL-7B&0.419& 0.527& 0.586& 0.649& 0.923& 0.092& 0.120& 0.064\\
 Qwen2.5-VL-32B&0.317& 0.754& 0.841& 0.447& 0.918& 0.183& 0.258& 0.107\\
 Qwen2.5-VL-72B& 0.575& 0.744& 0.826& 0.686& 0.918& 0.266& 0.345& 0.187\\
 \midrule
 Ferret-UI2 (Gemma-2B)& 1& 0.789& -& -& 0.715 & 0.070& 0.108& 0.032\\
 Ferret-UI2 (Llama3-8B)& 1& 0.758& -& -& 0.855 & 0.083& 0.119& 0.046\\
 OmniParser V2& -& -& -& -& -& 0.290& 0.382& 0.198\\

 \midrule

    UI-UG-7B-SFT$_\text{{L}}$& 0.998 & 0.975 & 0.836 & 0.874 & 1 & 0.489 & 0.508 & 0.47 \\
   UI-UG-7B-SFT$_\text{{A+L}}$ & 1 & 0.972 & 0.859 & 0.9 & 1 & 0.544 & 0.567 & 0.527 \\
  UI-UG-7B-SFT$_\text{{V+A+L}}$ & 0.998 & 0.972 & 0.862 & 0.906 & 1 & 0.451 & 0.466 & 0.435 \\
   UI-UG-7B-SFT$_\text{{L}} $(\text{U}$_\text{{Only}}$)& 1 & 0.973 & 0.894 & 0.935 & 1 & 0.538 & 0.561 & 0.516 \\
   UI-UG-7B-SFT$_\text{{A+L}}$ (\text{U}$_\text{{Only}}$)& 1 & 0.968 & 0.892 & 0.932 & 1 & 0.527 & 0.547 & 0.507 \\

   UI-UG-7B-SFT$_\text{{V+A+L}} $(\text{U}$_\text{{Only}}$)& 1 & 0.972 & 0.906 & 0.953 & 1 & 0.499 & 0.526 & 0.472 \\
  UI-UG-7B-SFT$_\text{{A+L}}$\text{+GRPO}$_\text{U}$ & 1& 0.977 & 0.864 & 0.936 & 1 & 0.569 &0.598 & 0.552 \\

  UI-UG-7B-SFT$_\text{{A+L}}$\text{+DPO}$_\text{G}$ & 1& 0.974& 0.855 & 0.935 & 1 & 0.535 &0.558 & 0.512 \\

 \midrule
  UI-UG-7B& 1& 0.974& 0.854 & 0.933 & 1 & 0.559 &0.578 & 0.540 \\
\bottomrule
\end{tabular}
\caption{Performance comparison of different models on referring and grounding tasks.}
\label{tab1}
\end{table*}

\begin{quote}
\begin{small}

\begin{figure*}[t]
\centering
% \hspace*{-0.3cm}
\includegraphics[width=2.0\columnwidth,page=1]{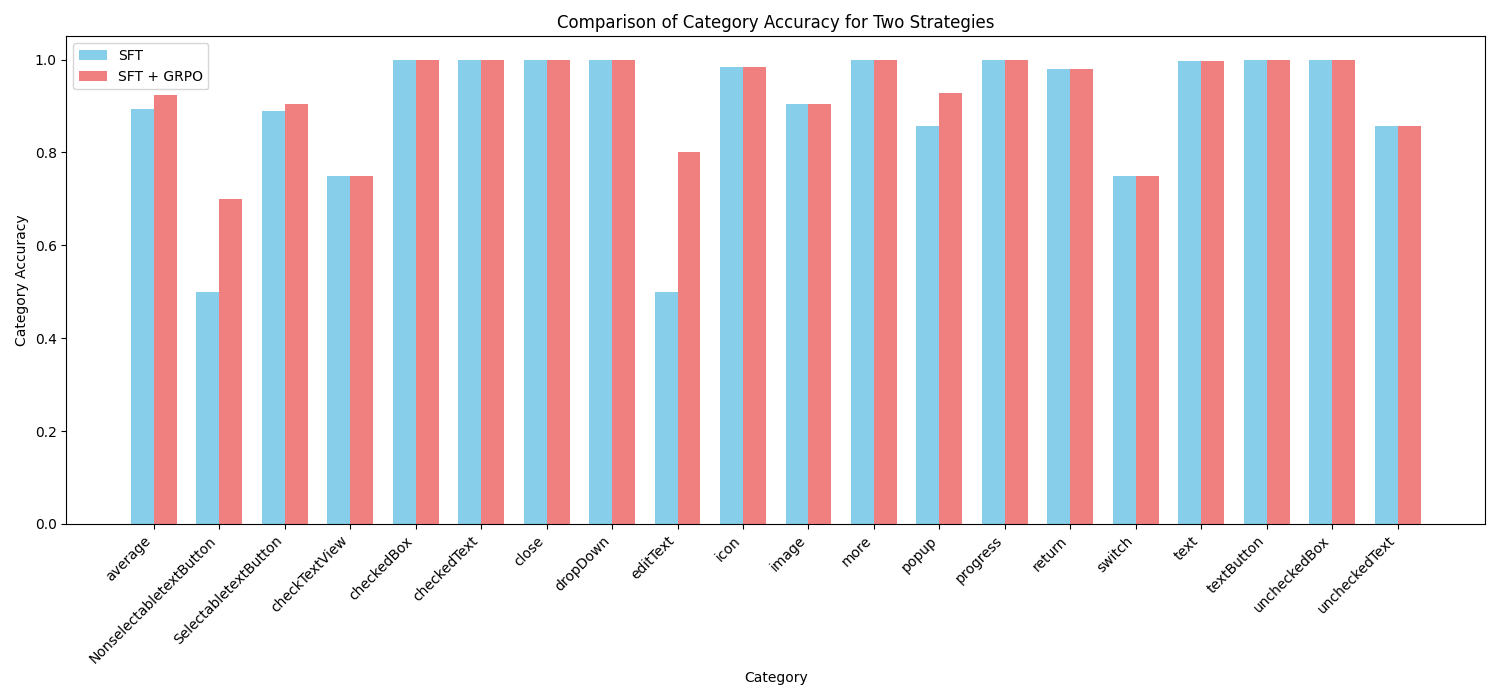}
\caption{Improvements of performance for referring tasks for each category by reinforcement learning. }
\label{figgrpo}
\end{figure*}

prompt\_templates\_grounding = [ \\
        "List all the \{category\} items.", \\
        "Please list all the \{category\} in the image.", \\
        "Identify every \{category\} present in this picture.", \\
        "List all \{category\} shown in the image.", \\
        "Enumerate each \{category\} in the provided photo.", \\
        "What are all the \{category\} in this image? Please enumerate.", \\
        "Provide all \{category\} types found in this picture.", \\
        "Can you list every \{category\} appearing in this image?", \\
        "Give a detailed list of every \{category\} in the picture.", \\
        "Please identify all of the \{category\} in the image.", \\
        "List every type and position of \{category\} displayed in the image." \\
] \\
prompt\_templates\_referring = [ \\
    "Describe the region \{box\}.", \\
    "Describe the categories and properties of the UI in area \{box\}.", \\
    "List the types and characteristics of UI elements in region \{box\}.", \\
    "Provide details on the UI types and their attributes within area \{box\}.", \\
    "Explain the categories and features of the UI components in region \{box\}.", \\
    "Summarize the types and properties of UI present in area \{box\}." \\
] \\
prompt\_templates\_ocr = [ \\
        "Describe the text present in area \{box\}.", \\
        "List all the textual information in area \{box\}.", \\
        "Please detail any text found in area \{box\}.", \\
        "What text appears in region \{box\}? Please explain.", \\
        "Summarize the main textual content in region \{box\}." \\
] \\

prompt\_templates\_text\_color = [ \\
    "Describe the text colors in region \{box\}.", \\
    "List all text colors found in area \{box\}.", \\
    "What text colors are used in region \{box\}? ", \\
    "Summarize the colors of the text in region \{box\}.", \\
    "Please state the colors of various texts in area \{box\}." \\
] \\

prompt\_templates\_caption = [ \\
        "Describe this UI image.", \\
        "Please describe this UI image.", \\
        "Introduce the content of this UI image.", \\
        "Explain in detail the content displayed in this UI image.", \\
        "Please analyze the content of this UI image." \\
] \\

prompt\_templates\_generation = [ \\
    "Design a UI card to showcase the details of \{ui\_description\}", \\
    "Design a concise, \\ clear, \\ and visually appealing UI card to display\{ui\_description\}", \\
    "Design a UI card to display \{ui\_description\}", \\
    "Please design a UI card displaying \{ui\_description\}", \\
    "Please generate a UI card to guide users to the \{ui\_description\}", \\
    "Design a UI card to showcase the details of \{ui\_description\}. The following is the mock data for the UI card.\{mock\_data\}", \\
    "Design a concise, \\ clear, \\ and visually appealing UI card to display\{ui\_description\}. The following is the mock data for the UI card.\{mock\_data\}", \\
    "Design a UI card to display \{ui\_description\}. The following is the mock data for the UI card.\{mock\_data\}", \\
    "Please design a UI card displaying \{ui\_description\}. The following is the mock data for the UI card.\{mock\_data\}", \\
    "Please generate a UI card to guide users to the \{ui\_description\}. The following is the mock data for the UI card.\{mock\_data\}" \\
] \\

\end{small}
\end{quote}

\subsubsection{Prompts for General-purpose MLLMs' UI Understanding Evaluation.}

\begin{quote}
\begin{small}
cat\_str = "text, image, SelectabletextButton, close, popup, icon, checkedText, NonselectabletextButton, uncheckedText, uncheckedBox, checkTextView, switch, progress, textButton, dropDown, editText, return, more, checkedBox"

\textbf{For referring.} \\
Your task is to give description of the given region in the image from four perspectives. First, you should decide the category of the given region, the reference categories are "{cat\_str}". Second, you should describe the text content in the given region. Third, you should output the color the text you extract in the second step in the hexadecimal representation such as \#414141.

You should output the answer in json format like: {"category": "the category you decide", "text": "the text you extract", "text\_color": ["the color you decide"]} and do not return any other content.

\textbf{For grounding.} \\
Your task is to detect all the given ui elements in the image. First, you should output the category of all the ui element you detect, you can refer to the following categories: "{cat\_str}". Second, you should output the bounding box of all the ui element you detect in the format of [x\_min, y\_min, x\_max, y\_min].' 

Please return the answer in json format like: [\{"type": \{ui element\}, "box": [x\_min,y\_min,x\_max,y\_max]"\},...]. \\\\
(optional: Scale the coordinates between 0 and 1.)
\\ **Note**: Please do not create new ui elements and do not return any other content.' 

\end{small}
\end{quote}

\begin{figure*}[t]
\centering
% \hspace*{-0.3cm}
\includegraphics[width=2.0\columnwidth,page=2]{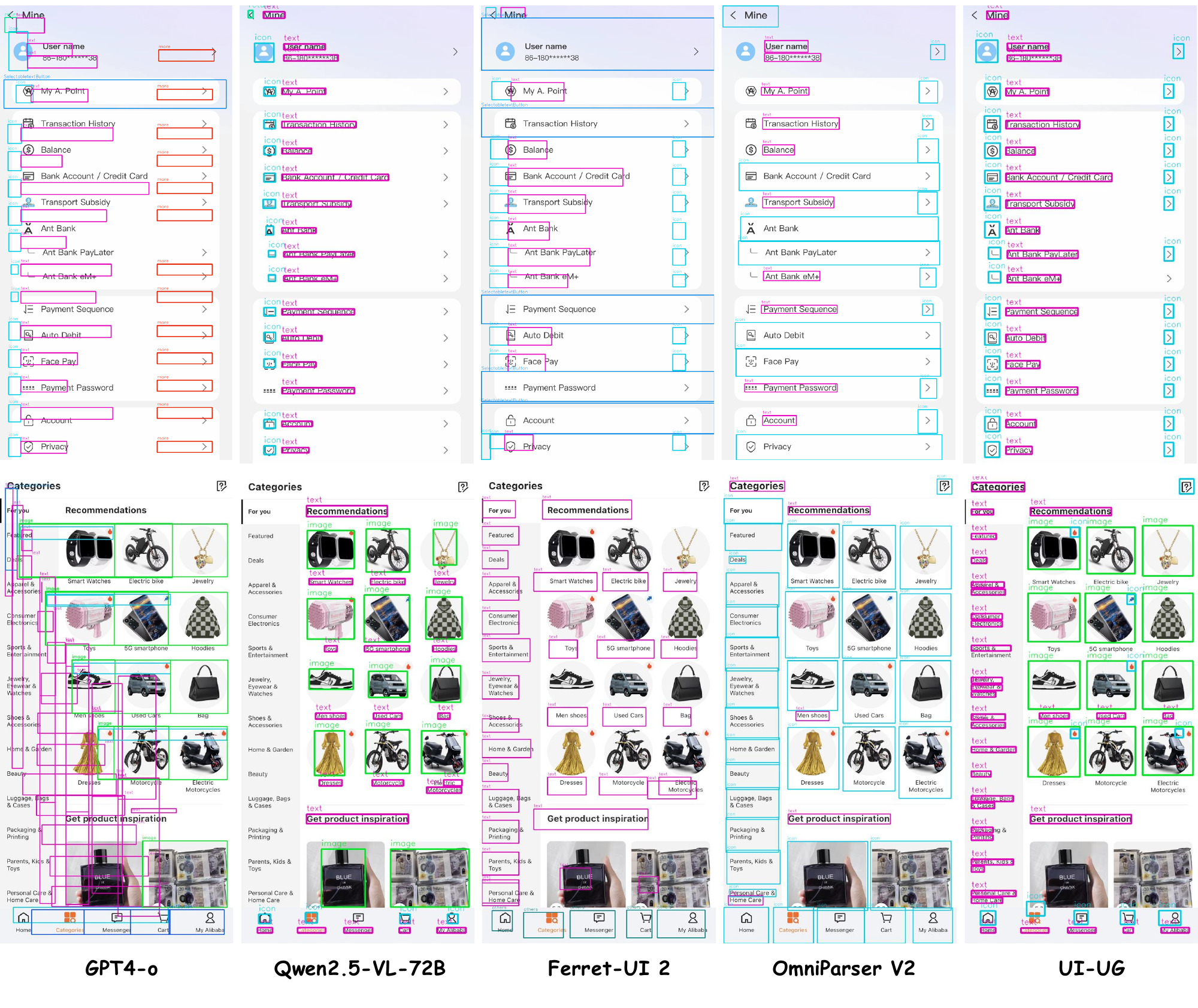}
\caption{Visual comparison of different models for grounding task for densely arranged UIs.}
\label{figgrounding1}
\end{figure*}

\begin{figure*}[t]
\centering
% \hspace*{-0.3cm}
\includegraphics[width=2.0\columnwidth,page=1]{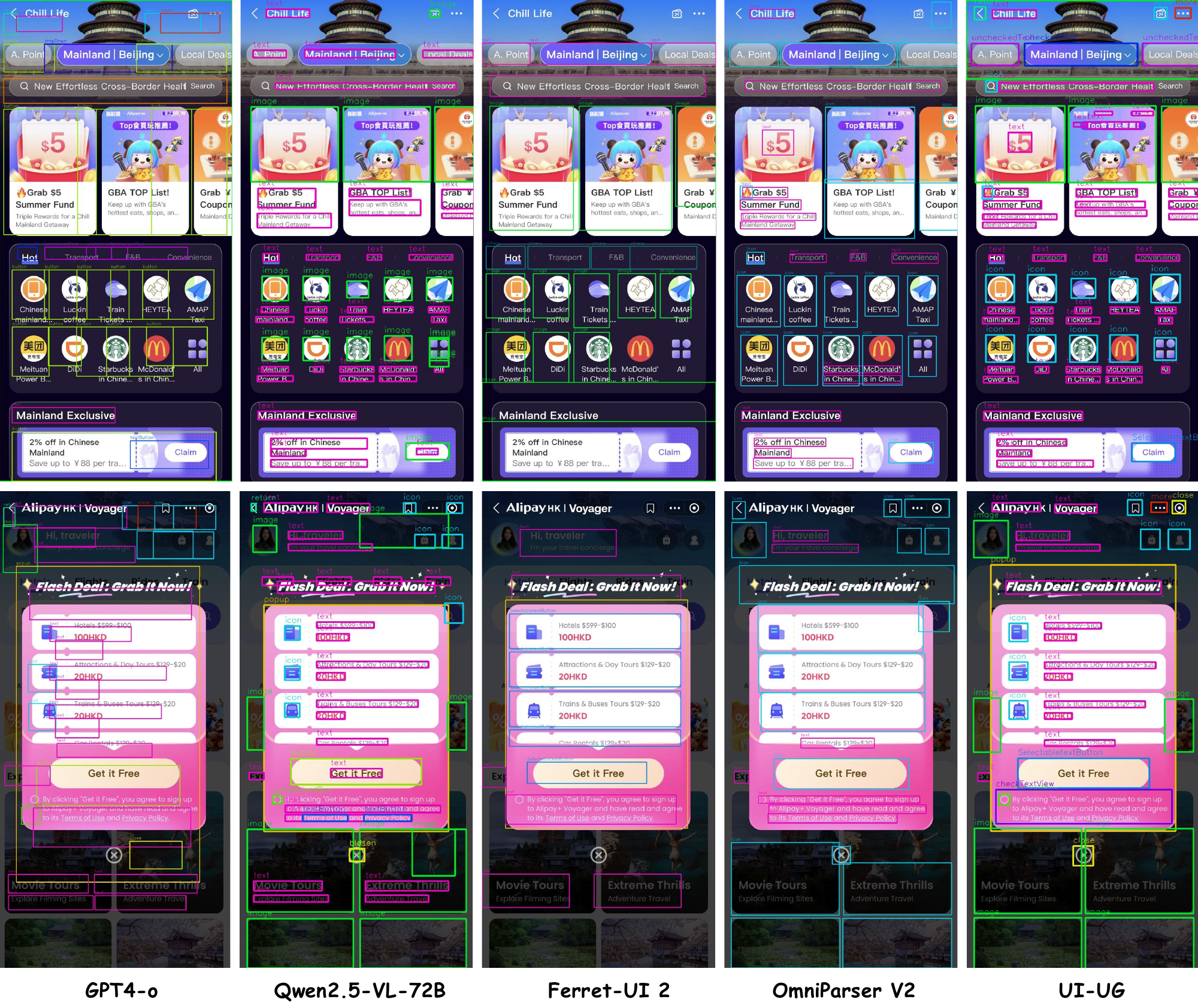}
\caption{Visual comparison of different models for grounding task for complex text-image mixed UIs. }
\label{figgrounding2}
\end{figure*}

\subsubsection{Prompts for Web2Code UI Generation Evaluation.}

\begin{quote}
\begin{small}
You are an advanced AI model equipped with OCR and image understanding capabilities, capable of analyzing visual elements in detail. 

Your task is to assess two UI images and output a score between 0 and 10 for each of the following questions.
If the answer to a question is a definite YES, output a score of 10, signifying perfect similarity. Conversely, a definite NO should yield a score of 0, indicating no similarity.
For answers that fall in between, assign a score accordingly, where a higher number indicates a greater degree of similarity. Only provide the numerical score for each question, without any additional text.
Example contexts are provided for clarity. Examples provides the idea, but you can output any number in 0-10 range accordingly. Only output a comma separated list containing 4 numbers. DO NOT give score of 10 for any category unless otherwise the two images are identical.

1. \textbf{Visual structure and alignment}(Score: 0-10):
Does the placement of elements like images, buttons, and text boxes aligned similarly on both images? 
Do the sizes and aspect ratios of images, buttons, and text boxes appear consistent across both images?  
Do both UI images exhibit a similar level of visual harmony and balance in their design? 
(e.g., A score of 10 for identical visual structure and alignment, 5 for similar but not exact consistent, and 0 for completely different structure.)

2. \textbf{Color match and Aesthetic Resemblance}(Score: 0-10):
How closely do the color schemes of the two UI images align in terms of background and text colors? Evaluate the similarity in hues, saturation, and overall color
aesthetics. Is the overall aesthetic appeal (modern, minimalistic, traditional, etc.) similar on both UI images? 
(e.g., A score of 10 for perfectly matching color schemes and identical aesthetics, including identical hues and saturation levels, 6 for similar color palettes and styles with minor variations, and 0 for starkly different color
schemes and aesthetics that create entirely different visual impacts.)

3. \textbf{Textual and Content Consistency}(Score: 0-10):
Do the font type, size, style, and weight of two UI images similar?
Do the words and sentences match between the two UI images?
(e.g., A score of 10 for complete uniformity in font type, size, style, weight across both UI images and identical text, 5 for consistency in font type and size but variations in style or weight, and 0 for wide disparities in font type, size, style, or weight, leading to a distinctly different textual appearance and content.) 

4. \textbf{User Interface Consistency} (Score: 0-10): 
Do the user interface elements (like menus, buttons, and forms) on both UI images share a similar design language and appearance? 
(e.g., A score of 10 for identical UI elements, 6 for slight design variations, and 0 for completely different UI designs.)

\end{small}
\end{quote}

\subsection{3. More Experiment Results}

\subsubsection{Results for UI Understanding Evaluation.}

Table \ref{tab1} presents the results of referring and grounding tasks across all the models, supplemented with performance comparisons from ablation experiments. Additionally, due to the imbalanced data distribution across different UI categories, the overall accuracy differences between models are not particularly significant. We conducted a detailed breakdown of subcategories within UI-UG (as shown in Figure \ref{figgrpo}), which demonstrates that our reinforcement learning approach indeed achieved improvements in challenging categories.

Notably, we observed \textbf{significantly lower mAP (mean Average Precision) values} in other models compared to ours. To investigate this issue, we visualized some sample images from our test set. Especially, Figure \ref{figgrounding1} and Figure \ref{figgrounding2} respectively compare typical models (including GPT4-o, Qwen2.5-VL-72B, Ferret-UI 2, and OmniParser V2) with our UI-UG model on densely arranged UIs and complex text-image mixed UIs. Our analysis confirms that our model exhibits superior classification accuracy and detection precision, particularly in:
\begin{itemize}
\item More accurate identification of UI categories.
\item Tighter bounding box predictions along elements.
\end{itemize}

\begin{figure*}[t]
\centering
% \hspace*{-0.3cm}
\includegraphics[width=2.0\columnwidth,page=3]{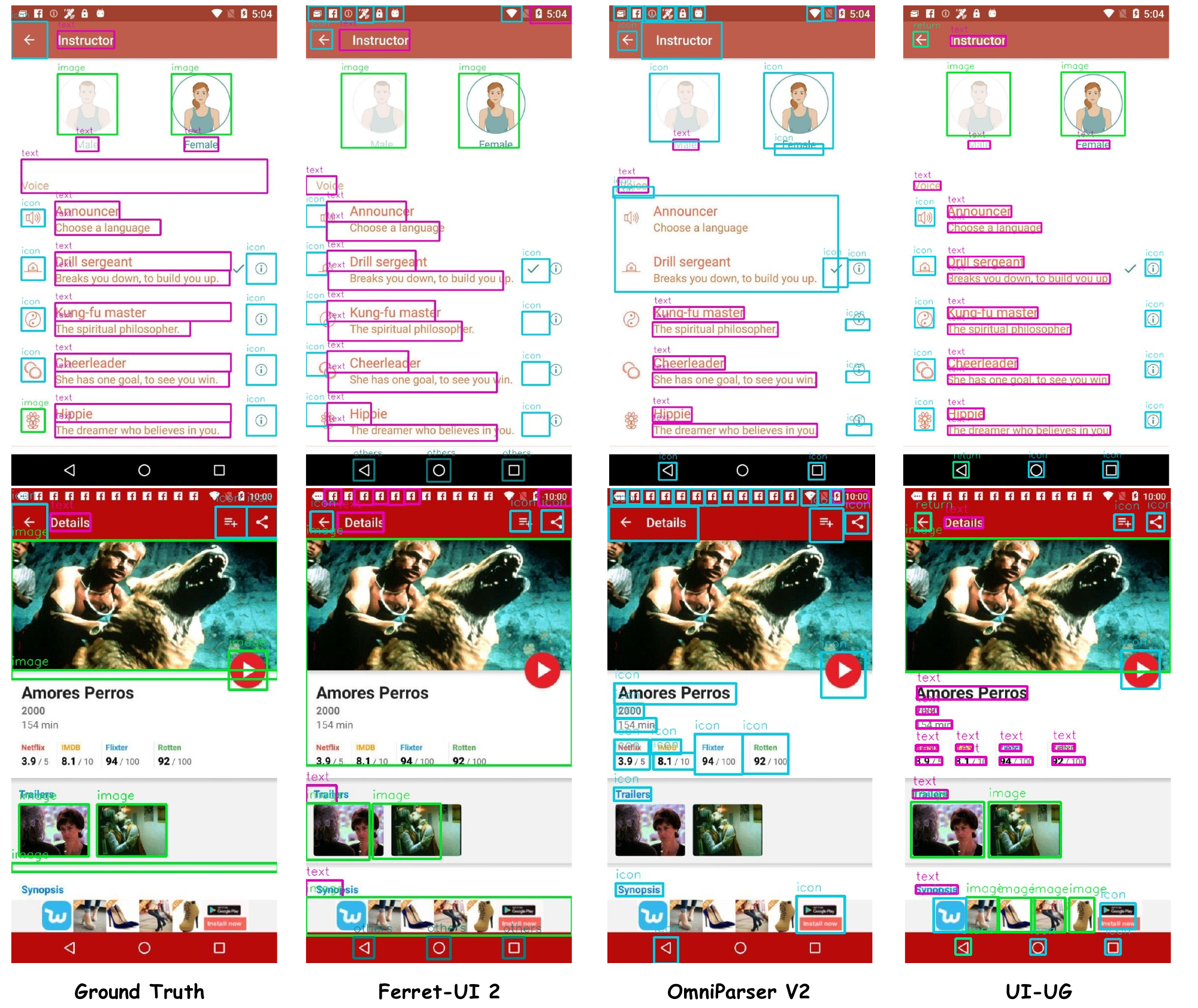}
\caption{Visual comparison of different models for grounding task in the RICO dataset.}
\label{figgroundingrico}
\end{figure*}

It should be emphasized that our final AP metric averages performance across \textbf{both simple and complex categories}. As well, given our focus on \textbf{small element detection} in UI images, the evaluation criteria  become inherently more stringent. These findings collectively validate that our evaluation outcomes align with theoretical expectations.

In addition, we mentioned in the paper that our model still outperforms other models on the zero-shot RICO dataset, although the overall mAP is lower compared to its performance on our primary dataset. Figure \ref{figgroundingrico} presents a visual comparison of grounding task results on the RICO dataset. We observe that our model's predictions demonstrate finer granularity compared to the ground-truth annotations in the RICO dataset, with bounding boxes showing better alignment. This visually substantiates the explanations provided in our paper.

\subsubsection{Results for UI Generation Evaluation.}

\begin{figure*}[t]
\centering
% \hspace*{-0.3cm}
\includegraphics[width=1.8\columnwidth]{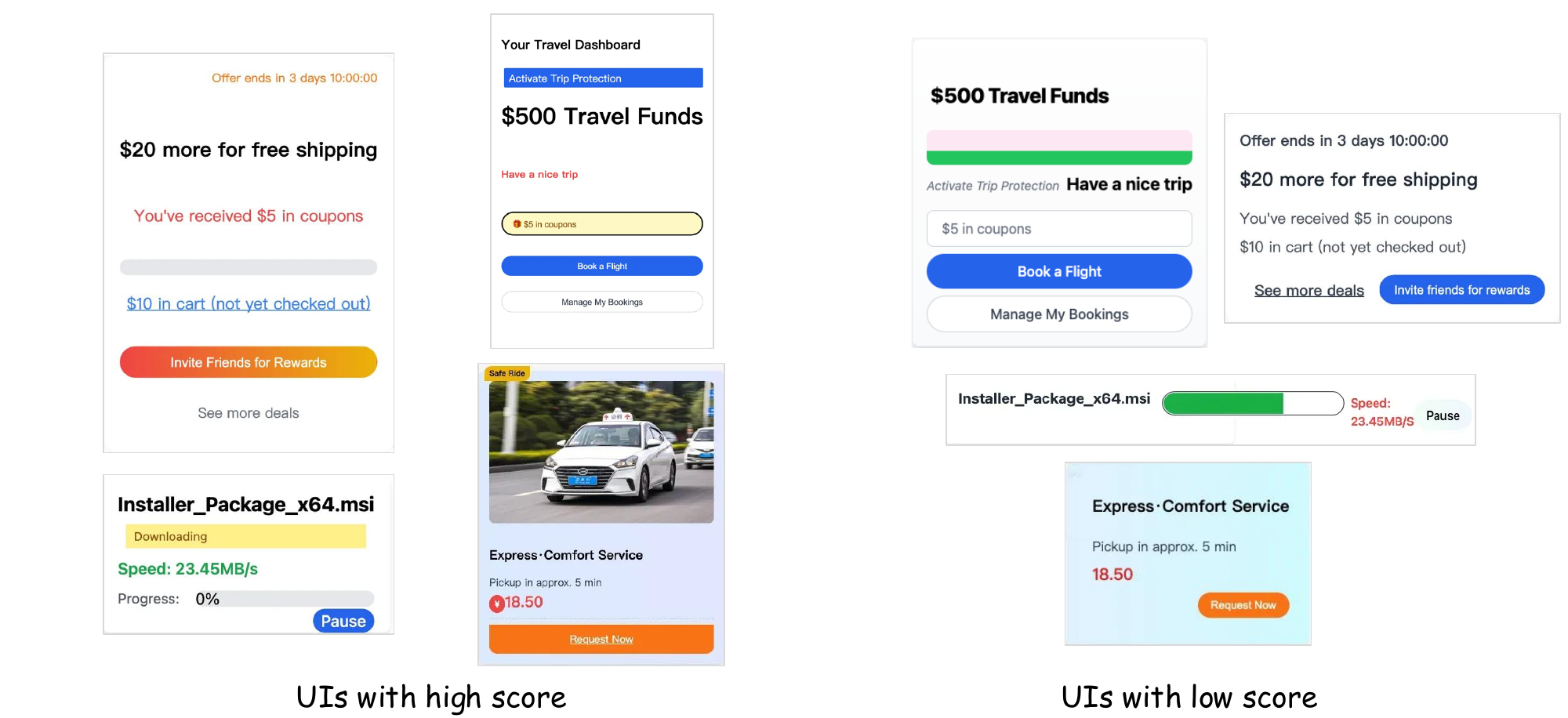}
\caption{Visual comparison of UIs with high score and low score (from DPO Dataset).}
\label{figdpo}
\end{figure*}

Figure \ref{figdpo} displays a preview of our DPO dataset, featuring comparisons between high-scoring and low-scoring UIs. Table \ref{tab2} presents the performance metrics of all general-purpose MLLMs (Multimodal Large Language Models) on the UI dataset, including detailed scores across various categories. Benefiting from Supervised Fine-tuning (SFT), our model achieves optimal performance in formatting tasks. Furthermore, through Direct Preference Optimization (DPO) training, it demonstrates significant improvements in visual evaluation metrics, reaching performance levels comparable to those of larger, more advanced general-purpose models.

\subsection{4. Code \& Demonstrations in the ZIP Package}
\subsubsection{Codes and Data.}
In addition to the technical contributions outlined in our paper, we have made available \textbf{training data samples} (folder: data/ ) and \textbf{model evaluation codes} (folder: code/ ) to support reproducibility and extended analysis.

\subsubsection{Real-time Rendering for UI generation.} 
For additional visual demonstrations and dynamic examples, please \textbf{refer to the video} (ui\_render\_demo.mp4 ) contained within the supplementary materials zip package. These multimedia resources provide extended illustrations of how to render UIs using our model.

\begin{table*}[t]
\centering
\begin{tabular}{l c c c c c c l}
\toprule
  & Format & Reference & Visual & Color \& & Textual& Interface &Total\\
 Model  & Accuracy & Similarity & Structure \&  & Aesthetic  & \& Content  & \& &\\
   &   &   & Alignment &  Design & \ Consistency & Interactivity&\\
\midrule
 GPT-4o& 8.80 & 8.97 & 6.36 & 6.59 & 6.82 &  6.22 &43.76 
\\
 Claude 3.7 Sonnet& 8.80 & 9.02 & 6.90 & 7.11 & 6.49 &  6.76 &45.08 
\\
 Gemini 2.5 Pro& 8.70 & 8.95 & 7.00 & 6.92 & 6.60 & 6.63 &44.80 
\\
 Qwen2.5-VL-72B& 8.10 & 8.92 & 6.28 & 6.63 & 6.16 & 6.06 &42.15 \\

 \midrule
 UI-UG-7B-SFT$_\text{{A+L}} $(\text{U}$_\text{{Only}}$)& 8.60 & 4.53 & 4.30 & 5.35 & 4.85 & 4.53 &32.16 
\\
    UI-UG-7B-SFT$_\text{{L}}$ & 8.60 & 8.87 & 4.34 & 5.28 & 5.12 & 4.44 &36.65 
\\
   UI-UG-7B-SFT$_\text{{A+L}}$ & 8.60 & 8.83 & 4.38 & 5.21 & 5.05 & 4.63 &36.70 
\\
  UI-UG-7B-SFT$_\text{{V+A+L}}$ & 8.60 & 8.90 & 4.52 & 5.47 & 4.80 & 4.65 &36.94 
\\

  UI-UG-7B-SFT$_\text{{A+L}}$\text{+GRPO}$_\text{U}$ & 9.10 & 8.86 & 4.64 & 5.33 & 4.77 & 4.64 &37.34 
\\

  UI-UG-7B-SFT$_\text{{A+L}}$\text{+DPO}$_\text{G}$ & 9.50 & 8.92 & 5.70 & 6.29 & 6.47 & 5.65 &42.53 \\

 \midrule
 UI-UG-7B& 9.30 & 8.91 & 5.71 & 6.17 & 6.28 &  5.65 &42.02\\

\bottomrule
\end{tabular}
\caption{Performance comparison of different models on generation tasks.}
\label{tab2}
\end{table*}

\end{document}